\documentclass[runningheads]{llncs}

\usepackage{eccv}

\usepackage{eccvabbrv}

\usepackage{graphicx}
\usepackage{booktabs}
\usepackage{comment}
\usepackage{fancybox}
\usepackage{tikz}
\usepackage{amsmath}
\usetikzlibrary{fit, arrows.meta,
	calc, 
	positioning}
\usepackage{array}
\def\enstrect{\textsc{Enstrect}}

\usepackage[accsupp]{axessibility}  %

\newcommand{\astfootnote}[1]{%
	\let\oldthefootnote=\thefootnote%
	\setcounter{footnote}{0}%
	\renewcommand{\thefootnote}{\fnsymbol{footnote}}%
	\footnote{#1}%
	\let\thefootnote=\oldthefootnote%
}

\usepackage{hyperref}

\usepackage{orcidlink}

\begin{document}

\title{\enstrect{}: A Stage-based Approach to \\2.5D Structural Damage Detection} 

\titlerunning{\enstrect{}: 2.5D Structural Damage Detection}

\author{Christian Benz\orcidlink{0000-0001-9915-005} \and
Volker Rodehorst\orcidlink{0000-0002-4815-0118}}

\authorrunning{C.~Benz \& V.~Rodehorst}

\institute{Bauhaus-Universität, Weimar, Germany
\\
\email{christian.benz@uni-weimar.de}}

\maketitle

\begin{abstract}
To effectively assess structural damage, it is essential to localize the instances of damage in the physical world of a civil structure. \enstrect{} is a stage-based approach designed to accomplish 2.5D structural damage detection. The method requires an image collection, the relative orientation, and a point cloud. Using these inputs, surface damages are segmented at the image level and then mapped into the point cloud space, resulting in a segmented point cloud. To enable further quantitative analyses, the segmented point cloud is transformed into measurable damage instances: cracks are extracted by contracting the clustered point cloud into a corresponding medial axis. For areal damages, such as spalling and corrosion, a procedure is proposed to compute the bounding polygon based on PCA and alpha shapes.
With a localization tolerance of 4\,cm, \enstrect{} can achieve IoUs of over 90\% for cracks, 82\% for corrosion, and 41\% for spalling. Detection at the instance level yields an $\text{AP}_{50}$ of about 45\% (cracks, spalling) and 56\% (corrosion)\astfootnote{Accepted Manuscript. To appear in the ECCV'24 workshop proceedings. Link to the published version on the publisher's website will be added as soon as available. Code can be found here \url{https://github.com/ben-z-original/enstrect}}. 
\end{abstract}

\section{Introduction}
\label{sec:intro}
The importance of structural health monitoring (SHM) for modern societies is undeniable. SHM ensures that critical civil infrastructure remains operational and, when necessary, is renewed in a timely and systematic manner. This contributes to the safe and enduring usage of vital infrastructure, which is essential for the functioning of contemporary societies. Transportation routes depend on reliable infrastructure for the delivery of goods and individual mobility. Bridges, in particular, are crucial for shortening routes and bypassing rough terrain, sometimes serving as the only viable connection between two locations. Exposure to weathering and other forces makes bridges especially vulnerable to degradation, and their unexpected collapse can have catastrophic consequences. The regular inspection of bridges is of significant societal benefit, however, it poses substantial challenges for both personnel and machines involved.

The availability and utilization of new technologies have spurred increased research activity in automating structural inspection. Notably, the advent of versatile imaging platforms, such as unmanned aircraft systems (UAS), commonly known as drones, and advanced data algorithms, like machine learning, have significantly advanced research and engineering in this field. With the continuous improvement and proliferation of technologies and data, the gap between research and real-world application is steadily closing.

As a result, research on image-based recognition of structural damage in critical infrastructure is rapidly expanding. While current efforts primarily focus on detecting cracks and other damages at the image level, translating this information into the 3D space of point clouds remains underexplored. The severity of damage significantly depends on its specific location on the structure, which is difficult to determine from images alone. Therefore, extending the detection of structural damage beyond the image level is the focus of this work.

To address this, a workflow is presented that achieves the detection of damage instances at the point cloud level. This workflow utilizes state-of-the-art models for structural damage segmentation at the image level and maps the 2D predictions onto the point cloud. However, the segmented point cloud lacks information about the damage instances, making further quantitative analyses -- such as measuring the length, width, or coverage of damages -- impossible. Therefore, the segmented point cloud undergoes further processing steps to transform the data into medial axes or bounding polygons, depending on the type of damage.

The main contributions of this work are: (1) the implementation of a fully functional pipeline for 2.5D damage detection, (2) the introduction of effective methods to transform the segmented point cloud into measurable damage instances, and (3) the integration of state-of-the-art image-level segmentation models into the pipeline, along with their evaluation on real-world data, a pioneering effort in this field\footnote{Dataset and model will be made publicly available upon publication.}.

\tikzset{procedure_style/.style={fill=gray!30, draw=gray!150, line width=1pt, rounded corners=2pt, minimum height=6mm}}
\tikzset{input_style/.style={fill=gray!5, draw=gray!150, line width=0.5pt, minimum height=6mm}}
\tikzset{stage_style/.style={draw=gray!150, line width=0.5pt, inner sep=10pt, rounded corners=5pt}}
\tikzset{img_style/.style={
		inner sep=0pt,     %
		outer sep=0pt,     %
		rectangle,
		align=center,
		yshift=-6mm
	}
}

\section{Related Work}
\subsection{Crack Segmentation}
Since 2017, artificial neural networks (ANN) have emerged as the dominant approach for crack detection. \cite{dorafshan2018comparison} conducted a study comparing different training configurations of AlexNet \cite{krizhevsky2012imagenet} with six edge detectors, including Sobel, LoG, and Butterworth. Experiments on the SDNET dataset \cite{dorafshan2018sdnet2018} indicated the superiority of ANN and demonstrated the effectiveness of transfer learning. Other approaches, proposed by \cite{zhang2016road, cha2017deep, chen2018nb} involve using a convolutional neural network (CNN) for classification combined with a sliding window to process larger images and/or improve localization. \cite{yang2018automatic} popularized the transition to fully-convolutional networks (FCN) \cite{long2015fully} for crack segmentation. Based on SegNet \cite{badrinarayanan2017segnet}, DeepCrack was designed \cite{zou2019deepcrack}: a separate fusion logic with individual scale-wise losses supports preserving thin structures. The conceptually similar and same-named approach DeepCrack is suggested by \cite{liu2019deepcrack}. In the style of deeply-supervised nets (DSN) \cite{lee2015deeply}, losses are computed for intermediate side-outputs to make use of fine details and anti-noise capabilities alike. The outputs undergo post-processing with guided filtering (GF) \cite{he2013guided} and conditional random fields (CRF) \cite{zheng2015conditional}. A U-Net \cite{ronneberger2015u} with focal loss \cite{lin2017focal} is reported to perform better compared to a simpler FCN design \cite{liu2019computer}. TernausNet \cite{iglovikov2018ternausnet}, a modified version of U-Net, is explored by \cite{benz2019crack} for detecting fine concrete cracks in images captured by an  UAS. \cite{yang2020feature} propose a feature pyramid and hierarchical boosting network, termed FPHBN. It extends holistically-nested edge detection (HED) \cite{xie2015holistically} by a feature pyramid module to incorporate and propagate context information to lower levels. The hierarchical boosting supports the interlevel communication within the FPHBN. 

\cite{liu2021crackformer} develop CrackFormer, which is a transformer-based approach to crack segmentation. For that purpose, the convolutional layers of VGG \cite{simonyan2015very} are replaced by a self-attention logic. To increase the crack sharpness, a scaling-attention block is suggested. \cite{benz2022image} propose a re-trained version of the hierarchical multi-scale attention network by \cite{tao2020hierarchical} called HMA, which mitigates the scale sensitivity of cracks. The results are aggregated based on the attention to cracks on different levels of scales. In order to preserve the continuity of cracks, \cite{pantoja2022topo} suggest TOPO loss, which uses maximin paths to mitigate discontinuities between cracks. An oriented bounding box approach, named CrackDet, has recently been proposed by \cite{chen2023the}. \cite{bianchi2022development} and \cite{kulkarni2022crackseg9k} emphasize the usefulness of transfer learning and compose smaller crack datasets into larger ones, Conglo, and CrackSeg9k. \cite{kulkarni2022crackseg9k} compare a number of approaches, including Pix2Pix, SWIN, and MaskRCNN. DeepLabV3+ \cite{chen2018encoder} with a ResNet-101 backbone outperformed the other methods. \cite{bianchi2022development} confirm that DeepLabV3+ is an effective method for crack segmentation. The extensive \textsc{OmniCrack30k} benchmark \cite{benz2024omnicrack30k} further indicates the effectiveness of nnU-Net for crack segmentation.

\subsection{Detection of Structural Damages}
Numerous datasets and methods for crack detection have been published. However, the focus on image-based detection of other structural damages has only recently gained traction. Similar to crack detection, the initial exploration for damage detection utilized the image classification paradigm. \cite{mundt2019meta} proposed a meta-learning approach for neural architecture search (NAS) on CODEBRIM, which slightly outperforms VGG-based \cite{simonyan2015very} and DenseNet-based \cite{huang2017densely} models while requiring significantly fewer parameters. \cite{flotzinger2022building} conducted benchmarking and extensive hyperparameter tuning for transfer learning on these datasets.

Since 2021, there has been a notable shift towards semantic segmentation for structural damage detection. \cite{benz2022image} introduced the structural defects dataset (S2DS), comprising 743 images featuring various types of damage, including cracks, spalling, corrosion, efflorescence, and vegetation. Their proposed model, DetectionHMA, leverages attention maps over different scales to effectively utilize multi-scale information.
To organize the growing number of datasets, \cite{bianchi2022visual} conducted a survey of published datasets for structural inspection and initiated a project to catalog available datasets\footnote{\url{https://github.com/beric7/structural_inspection_main/tree/main/cataloged_review}, accessed Jul 29, 2024}. The CSSC dataset \cite{yang2017a} includes images of cracks and spalling, while CrSpEE \cite{bai2021detecting} covers these damages in real-world scenarios with significant distractors such as people and complex backgrounds.
Recently, \cite{flotzinger2024dacl10k} released the dacl10k dataset, which contains 10,000 images and includes damage classes such as cracks, spalling, rust, efflorescence, wet spots, rock pockets, and weathering. The corresponding challenge, hosted at WACV'24 \cite{flotzinger2024dacl}, demonstrated the effectiveness of transfer learning on state-of-the-art models for semantic segmentation, including ConvNeXt-Large \cite{liu2022a}, EVA-02-Large \cite{fang2024eva}, and Mask2Former \cite{cheng2022masked}, with extensive use of model ensembles.

\subsection{3D Damage Detection}
Interest in the image-based detection of cracks and other structural damages in 3D space has been steadily increasing. Due to the field being in its early stages, the lack of publicly available datasets hinders the effective training and evaluation of models. In the industrial context, the MVTec 3D-AD dataset was introduced \cite{bergmann2022the}. This dataset includes point clouds featuring scratches, holes, deformations, and other damages across ten industrially relevant categories such as cookies, carrots, dowels, and ropes.
\cite{bergmann2023anomaly} adapted an unsupervised student-teacher approach to 3D for the MVTec 3D-AD dataset, inferring damaged point clouds from deep feature descriptors. However, the application context of MVTec 3D-AD differs from structural inspection due to the controlled acquisition circumstances, the higher 3D resolution achieved, and the nature of the objects represented in the dataset.

Specifically targeting 3D structural inspection, \cite{jahanshahi2012adaptive} proposed a system for 3D crack detection by combining information from multiple images using depth data obtained from structure-from-motion (SfM). \cite{torok2014image} developed a method to detect cracks in a triangulated mesh by analyzing the deviation of normals relative to the medial axis of an element. \cite{huang2014a} utilized depth information captured by a laser scanner to enhance image-based crack segmentation.
\cite{zhang2017automated} introduced CrackNet, a CNN designed to operate on depth maps for detecting pavement cracks. Building on this work, a more advanced learning-based model was later proposed by \cite{zhang2018deep}. \cite{chen2022crackembed} extracted embedding features for 3D points to segment crack regions in an unsupervised manner. \cite{pantoja2023damage} pre-segmented building facades and then projected cracks, segmented at the image level, onto the 3D model. A transformer-based method for controlling attention across different views in 3D crack detection is proposed by \cite{benz2024mvcrackvit}.

Given the comparatively low resolution of point clouds in real inspection scenarios, methods operating natively at the point cloud level are not particularly suitable for this application. Therefore, this work approaches damage detection beyond 2D by utilizing higher resolution data from multiple views. Due to the current scarcity of data, end-to-end training of a multi-view point cloud segmentation method is not yet feasible. Instead, this study leverages the effective performance of 2D damage detection in a multi-stage approach, transforming 2D results into 2.5D damage instances.

\section{Dataset}
Given the nascent nature of the field, no publicly available, properly labeled 2.5D or 3D datasets for structural damage detection exist. This absence can be attributed to several factors: the challenging accessibility of specific structures (such as bridges), the stringent quality standards required for the data, and the significant effort and expertise needed for labeling. Therefore, a key contribution of this work is the acquisition and preparation of a dataset representative to this particular use case.

\def\ind{\hspace{2ex}}
\begin{table}[t]
	\centering
	\caption{Dataset for 2.5D damage detection in a real-world inspection scenario split into a development and test sets. Two bridges with two segments each are used, featuring cracks, spalling, and corrosion.}
	\setlength{\tabcolsep}{5pt}
	\begin{tabular}{lcccc}
		\toprule
		& \multicolumn{2}{c}{Bridge B} & \multicolumn{2}{c}{Bridge G} \\
		& Dev & Test & Dev & Test \\\midrule
		\begin{tabular}{@{}l@{}}
			Mesh \\
			\ind \# Faces \\
			\ind Area [m$^2$] \\
			Views \\ 
			Damage Instances \\
			\ind Cracks \\
			\ind Spalling \\
			\ind Corrosion \\
		\end{tabular} & 
		\begin{tabular}{@{}c@{}}
			\\
			226,400\\
			0.69\\
			20 \\
			\\
			37 \\
			-- \\
			-- 
		\end{tabular} & 
		\begin{tabular}{@{}c@{}}
			\\
			198,894\\
			0.60\\
			22 \\
			\\
			20 \\
			-- \\
			--			
		\end{tabular} & 
		\begin{tabular}{@{}c@{}}
			\\
			64,862\\
			0.99\\
			28 \\
			\\
			-- \\
			9 \\
			8	
		\end{tabular} & 	
		\begin{tabular}{@{}c@{}}
			\\
			129,240\\
			1.97\\
			16 \\
			\\
			-- \\
			8 \\
			13
		\end{tabular}
		\\
		\bottomrule
	\end{tabular}
	\label{tab:data}
\end{table}

\cref{tab:data} provides an overview of the created dataset. The 3D reconstructions were obtained using SfM, multi-view stereo (MVS), and surface reconstruction techniques \cite{schonberger2016structure, schonberger2016pixelwise}. The points are sampled over the textured mesh. Bridge\,B is a highway bridge with comparatively wide cracks ranging from 0.5 to 1.0 mm. The accessible pier of this bridge was captured using a Sony $\alpha$7R I camera, which was moved around the pier's base on a tripod adjusted to two different altitudes. The second structure, Bridge\,G, is a railway bridge exhibiting spalling and corrosion. The images were captured with a UAS Intel Falcon 8+, equipped with a Sony $\alpha$7R I camera mounted on a compatible gimbal. Four segments showing relevant damages at niches or corners were extracted from the two bridges. One segment from each bridge was used in the development set for parameter tuning, while the remaining segments formed the test set. For a visual impression of the two test segments, kindly refer to \cref{fig:quali}.

Since the resolution of the reconstructed dense point cloud is typically lower than that of the image level, accurately annotating cracks and damage boundaries on the dense cloud is nearly impossible. To address this, a triangulated mesh with a high-resolution texture, nearly matching the image-level resolution, is computed and used for annotation. 
Polylines for cracks and bounding polygons for corrosion and spalling were annotated using the polyline tracing function provided by CloudCompare\footnote{\url{https://www.danielgm.net/cc/}, accessed Jul 16, 2024.}.

\section{\enstrect}
\begin{figure*}[t!]
	\centering
	{\includegraphics[width=\textwidth, trim=0ex 0 0 0, clip]{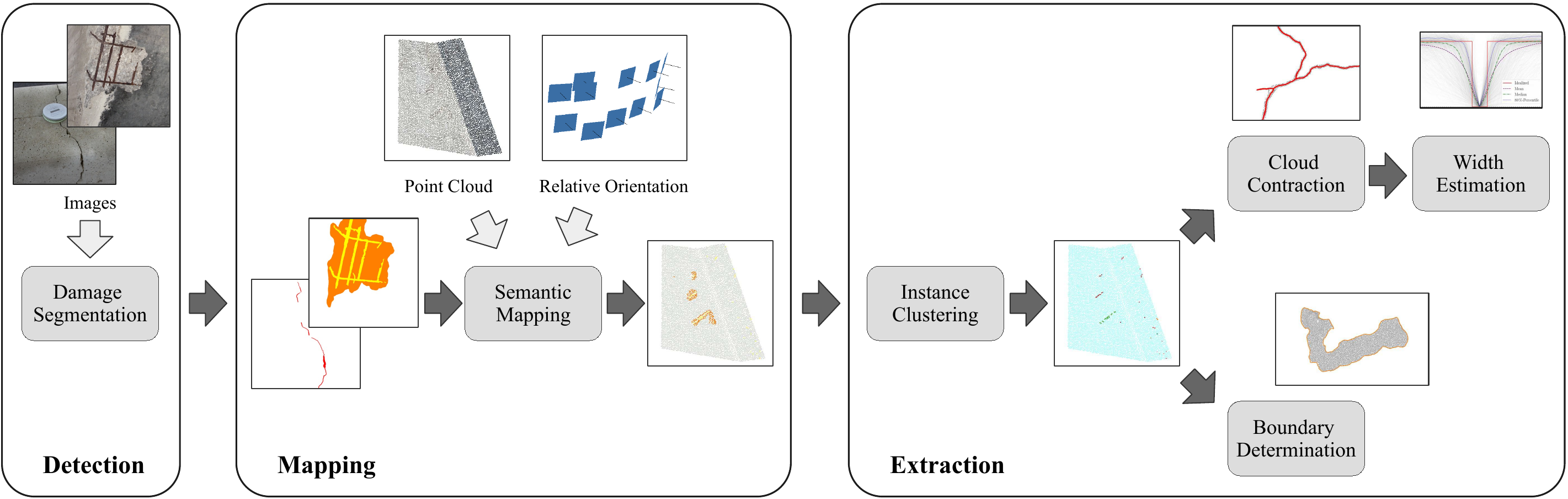}}
	\caption{Components and workflow of the 2.5D detection pipeline.}
	\label{fig:workflow}
\end{figure*}

\enstrect{} stands for \textit{\underline{en}hanced \underline{str}uctural insp\underline{ect}ion}. \cref{fig:workflow} schematically depicts the components of the workflow from 2D images into 2.5D damage instances. 3D reconstruction is assumed to have been performed successfully, ensuring that relative orientation and the point cloud are available. The three major stages of \enstrect{} are (1) \textit{detection} (``Damage Segmentation''), (2) \textit{mapping} (``Semantic Mapping'') and (3) \textit{extraction} (``Instance Clustering'', ``Cloud Contraction'', ``Width Estimation'', and ``Boundary Determination'').

\subsection{Detection}

Off-the-shelf models for image-level damage segmentation can be utilized. These models process the color images and generate probability maps, or \textit{heatmaps}, for each relevant class (\textit{crack}, \textit{spalling}, \textit{corrosion}, and \textit{background}). In this work, three state-of-the-art approaches—TopoCrack \cite{pantoja2022topo}, nnU-Net \cite{isensee2021nnu}, DetectionHMA \cite{benz2022image}—all based on CNNs, are used and compared.

\subsubsection{TopoCrack} With the goal of preserving the crack continuity, \cite{pantoja2022topo} introduce a novel topological loss called \textit{TOPO loss} and benchmark it against other losses. The base architecture is formed by TernausNet \cite{iglovikov2018ternausnet}, which is a U-Net-based architecture with a VGG11 encoder. Besides the dice loss, MSE for distance regression and TOPO loss for topology preservation are explored. Distance regression is based on truncated distance maps, which are inferred from the segmentation labels. Each pixel in the distance map represents its distance to the closest crack. Distances over 20 pixels are truncated. The truncated distance maps are used with MSE loss to enforce the model to learn the correct distances to the closest crack. 
The proposed combination of MSE and TOPO loss achieves an F$_1$ score of 69\% on the accompanying dataset.

\subsubsection{nnU-Net} 
The nnU-Net approach was proposed by \cite{isensee2021nnu} and published in Nature Methods in 2021. It achieved outstanding performance on several challenges in the medical domain. The main benefits of nnU-Net are its self-configuration properties, which require minimal manual intervention for designing strong models for 2D and 3D segmentation tasks alike. 
The term nnU-Net refers to `no new net' since it does not propose any new network architecture, loss function, or training scheme \cite{isensee2021nnu}. The process of `methods configuration' is systematized and delegated to a set of fixed, rule-based, and empirical parameters for automated self-configuration. Using the \textit{data fingerprint} derived from the specific dataset, heuristic rules guide the \textit{rule-based parameters} for data handling, such as resampling strategies, intensity normalization, patch and batch sizes, and the adjustment of the U-Net-based architecture template. Training is conducted using \textit{fixed parameters} for the optimizer, learning rate, data augmentation, and loss function, following a 5-fold \textit{cross-validation} scheme. In the post-processing step, an \textit{ensemble} is empirically determined. The nnU-Net was trained on the S2DS dataset (detailed below).

\subsubsection{DetectionHMA}
DetectionHMA, proposed by \cite{benz2022image}, is designed for detecting structural damages on concrete surfaces. It builds upon the hierarchical multi-scale attention (HMA) approach introduced by \cite{tao2020hierarchical}. To address scale-invariance issues often seen in CNNs, HMA incorporates multiple scales and dynamically combines results from different scales based on concurrently generated attention maps. These attention maps are contrastively learned using only two scales, though the number of scales used during inference can be chosen arbitrarily.
The backbone of HMA is the HRNet-OCR \cite{yuan2020object}, where OCR stands for object-contextual representations. These representations enhance pixel representation with contextual information. HMA employs the \textit{region mutual information} (RMI) loss, introduced by \cite{zhao2019region}, which combines a cross-entropy component with a mutual information component.

DetectionHMA was trained on the S2DS dataset, which comprises 743 images \cite{benz2022image}. This dataset contains images from real inspection sites taken with various camera types and includes the classes {background}, {crack}, {spalling}, {corrosion}, {efflorescence}, {vegetation}, and {control point}. However, due to the lack of 2.5D data for some classes, this work focuses on the detection of cracks, spalling, corrosion, and background.

\subsection{Mapping}
\label{sec:mapping}
In the mapping stage, image-level results are back-projected onto the point cloud. Points are projected into all views, and the collected information is aggregated to assign a single class label to each 3D point. 
Based on the assumption that views more perpendicular to a point provide higher certainty about a point's class membership, views from perpendicular angles are meant to contribute more than those from oblique directions. The deviation between the view and the point is measured by the angular difference between the point's normal and the viewing direction. The following weighting scheme has proven effective for this use case. Its major advantage lies in its simplicity, making it highly comprehensible and explainable for experts in the field of structural inspection:
\begin{align}
	w_i = 
	\begin{cases}
		\frac{1}{N}, &\text{if } 130^\circ < \theta < 230^\circ\\
		0, &\text{else}
	\end{cases}
	\label{eq:weighting}
\end{align}
The term $w_i$ refers to the weight for view $i$ for a certain point of the dense point cloud. $N$ denotes the number of views the point is visible in. The view is only considered if the angular deviation $\theta$ between the viewing direction and the point normal is in the range $(130^\circ, 230^\circ)$. Views outside this range are neglected. Weighting is performed on each class channel individually. Subsequently, a class label is assigned to the point based on an argmax/winner-takes-all logic, i.e. the class with the highest value is assigned and the others are discarded.

\subsection{Extraction}
The mapping process results in a segmented point cloud, consisting of independent points with no knowledge of their neighborhood. Consequently, it remains unclear which points represent the same or different instances of damage. For advanced quantitative analyses, such as measuring the length, width, or coverage of damage, it is essential to extract discrete instances of damage. The transformation process from a segmented point cloud into distinct damage instances is one of the major contributions of this work.

Before extracting instances, the segmented point cloud must be grouped into clusters that represent individual instances. Points of the same class and in close proximity are assumed to represent the same damage instance. The \textit{density-based spatial clustering of applications with noise} (DBSCAN) algorithm \cite{ester1996a}, which groups points based on their distance, is used for this purpose. A local high-density neighborhood, representing a cluster, is formed when a specified minimum number of points are within a certain distance. Low-density neighborhoods are considered noise and are consequently discarded.

\subsubsection{Cracks}
The point cloud representation is not suitable for quantitative analyses of detected cracks, such as determining their length, number of branches, or direction of propagation. To facilitate these analyses, the clustered point cloud is transformed into a polyline representation known as the (curve) skeleton or \textit{medial axis}. Although there are slight differences between these terms, they are used interchangeably in this work. \cref{fig:extraction} (a) to \cref{fig:extraction} (d) illustrate the stages of crack extraction. 

\cref{fig:extraction} (a) shows the crack on a mesh with high-quality texture. Through clustering, we obtain a subcloud that represents the specific instance of the crack, as shown in \cref{fig:extraction} (b). To extract the medial axis of the subcloud, \textit{Laplacian-based contraction} \cite{cao2010point} is applied\footnote{The implementation is based on \url{https://github.com/meyerls/pc-skeletor}.}. 
Laplacian-based contraction minimizes the quadratic energy \cite{cao2010point}:
\begin{align}
	\arg\min_{P'} \left( ||W_L L P'||^2 + \sum_{i} W_{H,i}^2||p'_i - p_i||^2 \right)
	\label{eq:contraction}
\end{align}
$P = \{p_i\}$ represents the original point cloud, $P' = \{p'_i\}$ the contracted point cloud, $L$ is the Laplacian matrix, $W_L$ the contraction weight matrix, and $W_H$ the attraction weight matrix. The first term $||W_L L P'||^2$ represents the geometric details, which are subject to smoothing. The second term $\sum_{i} W_{H,i}^2||p'_i - p_i||^2$ preserves the geometric shape of the point cloud. The contraction and the attraction weights balance the tendency to collapse into one point (``contraction'') and to remain at the current location (``attraction''). \cref{eq:contraction} is solved iteratively, employing increasing contraction weights and updated attraction weights. This approach maintains the shape and prevents full collapse.

\def\custfac{0.24}
\def\custfactwo{0.179}
\begin{figure}[t!]
	\centering
	\begin{tabular}{cccc}
		\frame{\includegraphics[height=\custfactwo\textwidth]{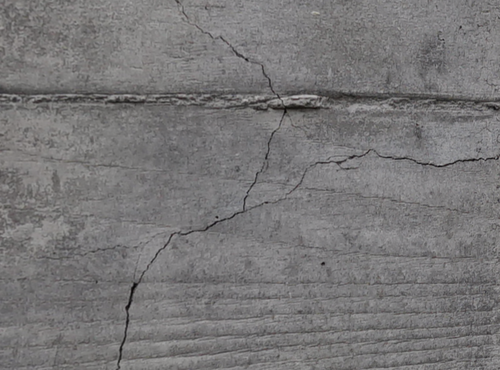}} &
		\frame{\includegraphics[height=\custfactwo\textwidth]{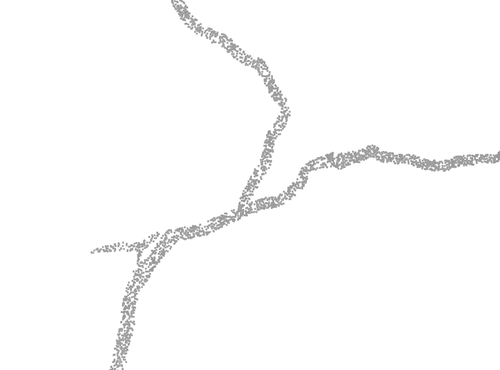}} &
		\frame{\includegraphics[height=\custfactwo\textwidth, clip]{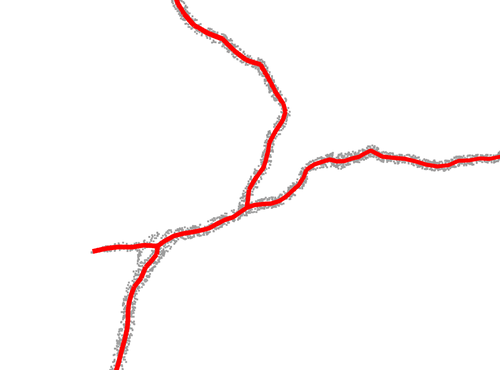}} &
		\frame{\includegraphics[height=\custfactwo\textwidth]{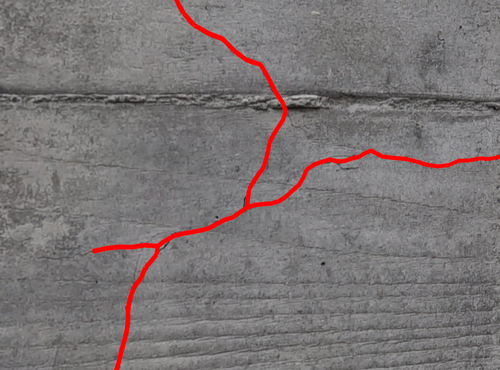}} \\
		(a)\,Textured mesh & (b)\,Clustered points & (c)\,Medial axis & (d)\,Axis on mesh
	\end{tabular}
	\caption{Illustration of the cloud contraction for extracting the medial axis of a branching crack: (a) shows a mesh with high-resolution texture, (b) the subcloud after clustering, (c) the overlayed medial axis (red) obtained by cloud contraction, and (d) the medial axis overlayed over the textured mesh.}
	\label{fig:extraction}
\end{figure}

The contracted point cloud lacks information about the connectivity of points. To address this, we propose the following procedure. First, a \textit{minimum spanning tree} is computed, resulting in a connected graph that includes all points with minimal edge lengths. The nodes of the spanning tree can have degrees of 1, 2, or more: nodes of degree 1 are end nodes, nodes of degree 2 are intermediate nodes, and nodes of degree 3 or more are branching nodes. The spanning tree is then recursively partitioned at each branching node to obtain branch-free polylines. \cref{fig:extraction} (c) displays the extracted medial axis (in red) for the respective point cloud, showing the detected crack as comprising five branch-free polylines. \cref{fig:extraction} (d) illustrates the medial axis overlaid on the textured mesh in 3D space. The extracted medial axis can, for instance, be used for the automated estimation of the crack width \cite{benz2021model}.

\subsubsection{Areal damages}
For areal damages such as spalling or corrosion, skeletonization is unsuitable. Bounding boxes and convex hulls, on the other hand, only provide a coarse outline of the damage's extent on the structure's surface. To better approximate the area covered by the damage, a \textit{bounding polygon} is computed. The proposed procedure is as follows: the subcloud representing the instance of spalling or corrosion is mapped into 2D space by performing principal component analysis (PCA) on the subcloud and retaining the two dimensions that explain the most variance. These dimensions are assumed to represent the plane in which the damage is located. Note that this procedure may fail for damages at corners, wall projections, and other not approximately planar regions.

\def\custfac{0.15}
\def\custfactwo{0.149}
\begin{figure}[t]
	\begin{tabular}{cccc}
		\centering
		\frame{\includegraphics[height=\custfactwo\textwidth]{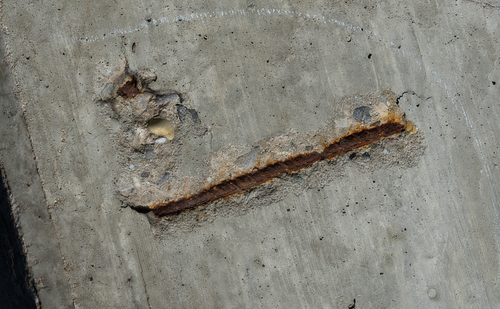}} &
		\frame{\includegraphics[height=\custfactwo\textwidth]{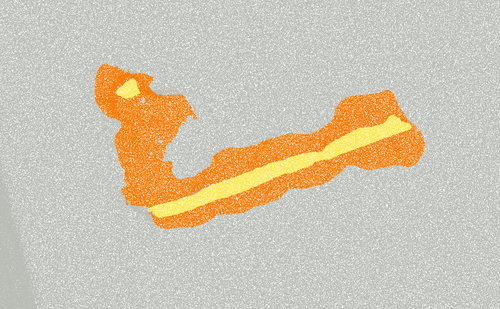}} &
		\includegraphics[height=\custfactwo\textwidth]{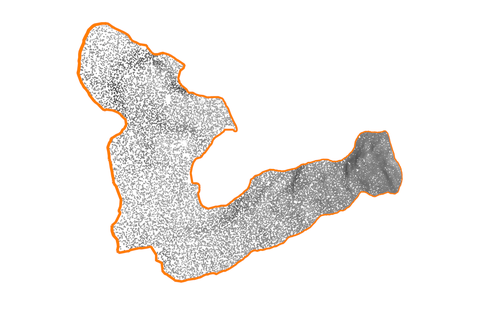} &
		\frame{\includegraphics[height=\custfactwo\textwidth]{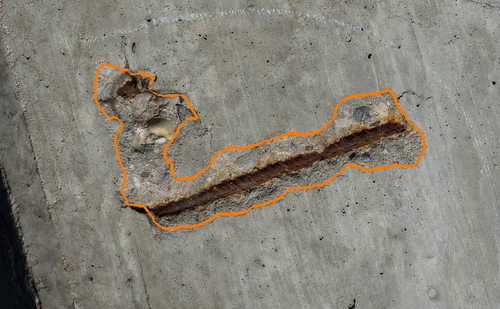}} \\
		(a)\,Textured mesh & (b)\,Segmented points & (c)\,2.5D polygon & (d)\,Polygon on mesh\\
	\end{tabular}
	\caption{Illustration of the extraction of the bounding polygon for areal damages: (a) shows a textured mesh with spalling and corrosion, (b) the segmented point cloud, (c) the corresponding bounding polygon in 2.5D space around the clustered point cloud (spalling and corrosion are merged in one cluster), and (d) the bounding polygon overlayed on the textured mesh.}
	\label{fig:alphashapes}
\end{figure}

The polygon extraction in 2D space is performed using alpha shapes \cite{edelsbrunner1983on, edelsbrunner1994three}. Alpha shapes generalize convex hulls and create a bounding alpha hull that encapsulates all relevant points. The parameter $\alpha$ represents the radius $1/\alpha$ of a generalized disk and controls the allowed concavity of the hull. For values of $\alpha$ close to zero, the alpha hull approximates the common convex hull. For $\alpha < 0$, the alpha hull is defined as ``the intersection of all closed complements of discs'' \cite{edelsbrunner1983on} with a radius of $-1/\alpha$. As $\alpha$ approaches negative infinity, the bounding hull with the highest concavity is obtained, corresponding to the minimum spanning tree of the points.

This work employs the implementation of alpha complexes\footnote{\url{https://github.com/bellockk/alphashape}, accessed Jul 29, 2024.}, which are closely related to alpha shapes. Instead of arcs, alpha complexes compute an alpha hull consisting of straight lines derived from Delaunay triangulation. The choice of $\alpha$ depends on the density of the points and the scale of the space. In this work, the PCA-transformed points are in normalized space, and an alpha value of $\alpha=100$ was found suitable. \cref{fig:alphashapes} (a) to \cref{fig:alphashapes} (d) illustrate the bounding polygon for an exposed reinforcement bar. \cref{fig:alphashapes} (a) shows the textured mesh of an exposed rebar, which in this work is modeled as the union of co-occurring spalling and corrosion. \cref{fig:alphashapes} (b) is a depiction of the segmented point cloud derived by the procedure described in \cref{sec:mapping}. Orange indicates spalling and yellow indicates corrosion. The bounding polygon is computed using alpha complexes in 2D space, in which the 3D subcloud was mapped using PCA. The vertices of the polygon in 2D space directly correspond to vertices in 3D, from which the 2.5D bounding polygon can be inferred, as shown in \cref{fig:alphashapes} (c). Finally, the bounding 2.5D polygon can be displayed alongside the textured mesh, \cref{fig:alphashapes} (d).

\section{Results}
\subsection{Evaluation Metrics}

\begin{table}[t!]
	\centering
	\caption{Quantitative test results for 2.5D damage detection. The approaches Topo\-Crack \cite{pantoja2022topo}, nnU-Net \cite{isensee2021nnu}, and DetectionHMA \cite{benz2022image} are compared. IoU refers to the intersection-over-union and AP$_{50}$ to the average precision (overlap 50\% or more) for instance-level evaluation. ``Tol.'' refers to the positional tolerance.}
	\begin{tabular}{@{}c@{}c|ccc|ccc|ccc}
		\toprule
		& Tol. & \multicolumn{3}{c|}{TopoCrack} & \multicolumn{3}{c|}{nnU-Net} & \multicolumn{3}{c}{DetectionHMA} \\
		& $[$cm$]$ & Crack & Spall. & Corr. & Crack & Spall. & Corr. & Crack & Spall. & Corr.\\
		\midrule
		\begin{tabular}{@{}l@{}}
			\rotatebox{90}{IoU $[$\%$]$} \\
		\end{tabular} & 					
		\begin{tabular}{@{}r@{}}
			1.0 \\ 2.0 \\ 4.0 \\ 6.0 \\ 8.0 \\
		\end{tabular} & 		
		\begin{tabular}{@{}r@{}}
			69.0 \\
72.7 \\
79.5 \\
83.3 \\
85.6\arraybackslash

		\end{tabular} & 
		\begin{tabular}{@{}r@{}}
			-- \\ -- \\ -- \\ -- \\ -- \\
		\end{tabular} & 
		\begin{tabular}{@{}r@{}}
			-- \\ -- \\ -- \\ -- \\ -- \\
		\end{tabular} & 
		\begin{tabular}{@{}r@{}}
			66.0 \\
71.3 \\
78.9 \\
85.0 \\
90.6\arraybackslash

		\end{tabular} & 
		\begin{tabular}{@{}r@{}}
			10.8 \\
17.4 \\
27.6 \\
43.7 \\
55.3\arraybackslash

		\end{tabular} & 
		\begin{tabular}{@{}r@{}}
			35.1 \\
75.8 \\
96.9 \\
99.5 \\
100.0\arraybackslash

		\end{tabular} & 	
		\begin{tabular}{@{}r@{}}
			89.0 \\
91.5 \\
94.9 \\
96.7 \\
99.0\arraybackslash	
		\end{tabular} & 
		\begin{tabular}{@{}r@{}}
			15.8 \\
25.4 \\
40.5 \\
58.3 \\
77.2\arraybackslash

		\end{tabular} & 
		\begin{tabular}{@{}r@{}}
			17.0 \\
47.0 \\
81.5 \\
89.7 \\
95.8\arraybackslash

		\end{tabular} \\	
		\midrule
		\begin{tabular}{@{}l@{}}
			\rotatebox{90}{AP$_{50}$ $[$\%$]$} \\
		\end{tabular} & 					
		\begin{tabular}{@{}r@{}}
			1.0 \\ 2.0 \\ 4.0 \\ 6.0 \\ 8.0 \\
		\end{tabular} & 		
		\begin{tabular}{@{}r@{}}
			5.6 \\
9.1 \\
14.9 \\
17.4 \\
22.2\arraybackslash

		\end{tabular} & 	
		\begin{tabular}{@{}r@{}}
			-- \\ -- \\ -- \\ -- \\ -- \\
		\end{tabular} & 
		\begin{tabular}{@{}r@{}}
			-- \\ -- \\ -- \\ -- \\ -- \\
		\end{tabular} &
		\begin{tabular}{@{}r@{}}
			8.6 \\
10.4 \\
17.5 \\
31.8 \\
44.8\arraybackslash

		\end{tabular} & 	
		\begin{tabular}{@{}r@{}}
			3.2 \\
16.2 \\
47.6 \\
61.7 \\
72.0\arraybackslash

		\end{tabular} & 
		\begin{tabular}{@{}r@{}}
			36.8 \\
68.8 \\
73.3 \\
78.6 \\
78.6\arraybackslash
	
		\end{tabular} &
		\begin{tabular}{@{}r@{}}
			22.3 \\
27.8 \\
45.0 \\
52.6 \\
62.5\arraybackslash

		\end{tabular} & 	
		\begin{tabular}{@{}r@{}}
			16.0 \\
32.7 \\
44.5 \\
53.3 \\
64.0\arraybackslash
	
		\end{tabular} & 
		\begin{tabular}{@{}r@{}}
			11.6 \\
49.0 \\
55.6 \\
55.6 \\
64.1\arraybackslash
	
		\end{tabular} \\	
		\bottomrule
	\end{tabular}
	\label{tab:results}
\end{table}

\begin{figure*}[t]
	\centering
	\def\triml{0ex} %
	\def\trimb{0ex} %
	\def\trimr{0ex} %
	\def\trimt{0ex} %
	\def\trimll{0ex} %
	\def\trimbb{0ex} %
	\def\trimrr{0ex} %
	\def\trimtt{0ex} %
	\def\trimrr{0ex} %
	
	\setlength{\tabcolsep}{2pt}
	\scriptsize
	\begin{tabular}{@{}cccc@{}}
	nnU-Net & nnU-Net\,\tiny{(zoomed)} & DetectionHMA & DetectionHMA\,\tiny{(zoomed)} \\
	{\includegraphics[width=0.24\textwidth, trim=10 \trimb{} \trimr{} \trimt{}, clip]{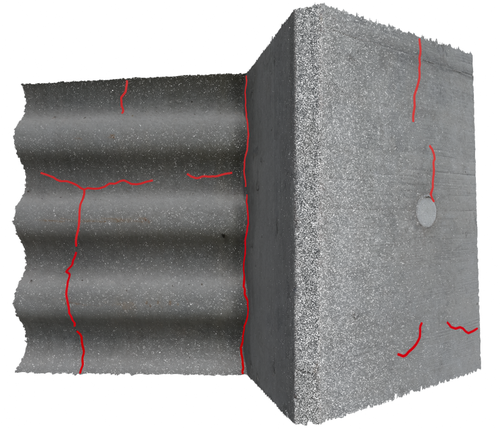}} &
	\frame{\includegraphics[width=0.24\textwidth, 
	clip]{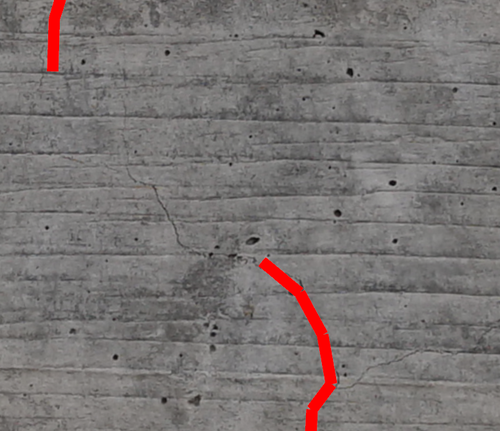}} &
	{\includegraphics[width=0.24\textwidth, trim=\triml{} \trimb{} \trimr{} \trimt{}, clip]{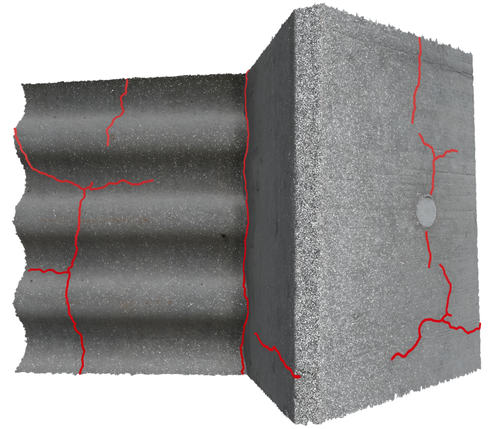}} &					
	\frame{\includegraphics[width=0.24\textwidth, 
	clip]{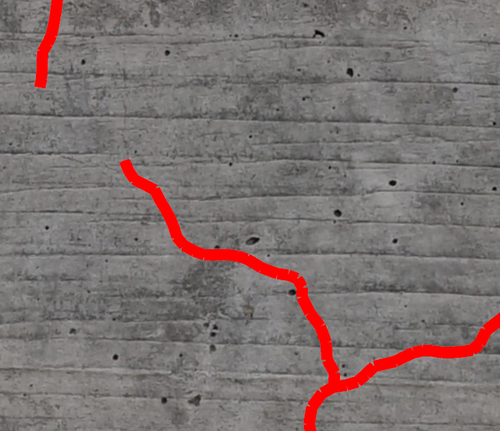}} \\
	{\includegraphics[width=0.24\textwidth, trim=20 5 \trimrr{} 5, clip]{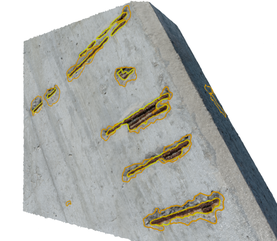}} &
	\frame{\includegraphics[width=0.24\textwidth, trim=\trimll{} \trimbb{} \trimrr{} \trimtt{}, clip]{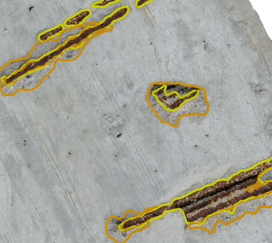}} &
	{\includegraphics[width=0.24\textwidth, trim=15 5 \trimrr{} \trimtt{}, clip]{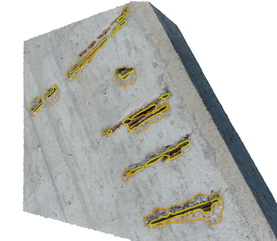}} &		
	\frame{\includegraphics[width=0.24\textwidth, trim=\trimll{} \trimbb{} \trimrr{} \trimtt{}, clip]{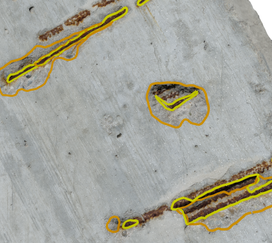}} \\		
	\end{tabular}
	\caption{Qualitative test results for 2.5D damage detection. The top row shows the test segment of Bridge\,B, the bottom row shows the test segment of Bridge\,G. The results of nnU-Net \cite{isensee2021nnu} and DetectionHMA \cite{benz2022image} are compared. While nnU-Net achieves better results for spalling and corrosion, DetectionHMA \cite{benz2022image} shows more robust performance for crack detection. In zoomed view: medial axis and bounding polygons overlayed on textured mesh. Best viewed on screen.}
	\label{fig:quali}
\end{figure*}

The basic evaluation procedure for the quantitative evaluation corresponds to \cite{knapitsch2017tanks,wang2018pixel2mesh,gkioxari2019mesh}. The vertices of both the medial axes and the bounding polygon are granted a positional tolerance $\tau$ based on the Euclidean distance measure $d$. The true positives (TP), the false negatives (FN), and the false positives (FP) are defined as: $\text{TP}(\tau) = \sum_{t\in \mathcal{T}}\left[d(t,p) \leq \tau\right]$, $\text{FN}(\tau) = \sum_{t\in \mathcal{T}}\left[\tau < d(t,p)\right]$, and $\text{FP}(\tau) = \sum_{p\in \mathcal{P}}\left[\tau < d(t,p)\right]$,
where $t$ denotes the true 3D vertices and $p$ the predicted 3D vertices of the medial axis resp.\ the bounding polygon. The square brackets $[\cdot]$ refer to the Iverson brackets, which evaluate to one when the respective conditional is fulfilled, zero otherwise. The \textit{intersection-over-union} is calculated by $\text{IoU}(\tau) = \text{TP}(\tau) / (\text{TP}(\tau) + \text{FN}(\tau) + \text{FP}(\tau))$.

To assess the instance detection capabilities of the proposed workflow, we use the standard metric of \textit{average precision} (AP), which is based on the integral of the precision-recall curve. An overlap threshold of $\text{IoU}(\tau)$ of 50\% is used; instances with equal or more overlap are considered true positives, while those with less overlap are classified as false positives or false negatives, respectively. Similar to IoU, $\text{AP}_{50}(\tau)$ is given a positional tolerance represented by the parameter $\tau$.

\subsection{Quantitative Results}
\cref{tab:results} presents the quantitative results. It is observed that TopoCrack performs reasonably well in terms of IoU but exhibits weaker performance in AP. This lower performance is likely due to the different dataset it was trained on: the TOPO dataset represents a different distribution than the S2DS dataset. The other two approaches were trained on the S2DS dataset. Since the TOPO dataset does not include classes beyond cracks, no results for spalling and corrosion are reported for TopoCrack.

DetectionHMA consistently outperforms the other two approaches in crack detection. For spalling, DetectionHMA leads in IoU, while nnU-Net achieves higher APs for larger tolerances. In corrosion detection, nnU-Net outperforms all other approaches across all tolerance levels. Performance consistently improves with increasing tolerance for all three approaches, as expected.

The strong performance of nnU-Net in detecting areal damages such as spalling and corrosion is particularly noteworthy and was reported in other work \cite{benz2023ai}. Its general-purpose design, which relies on self-configuration without manual intervention, yields results that are comparable to or even better than those of DetectionHMA, a model specifically fine-tuned for the domain dataset. However, nnU-Net falls short in crack detection. This discrepancy likely arises because many relevant classes in medical imaging more closely resemble blob-like objects rather than line-like objects in the few-pixels regime. This insight points to new potential directions for future research.

Although subject to debate, a positional tolerance of 4\,cm is assumed to offer a reasonable trade-off between positional accuracy and detection performance. At this tolerance level, the IoU values of DetectionHMA exceed 90\% for cracks and 80\% for corrosion, providing potential for effectively supporting structural inspections. However, spalling remains the most challenging category, achieving a top performance of 77\% only at an 8\,cm tolerance. It is hypothesized that detecting spalling solely from image features is difficult and would significantly benefit from a native 3D representation.

The instance detection numbers are notably lower, consistent with findings in other literature, where even slight offsets and discontinuities in object instances lead to reduced AP values. For more detailed automated quantitative analysis, AP values of 40\% to 50\% are inadequate. Exploring the underlying causes and implications of this lower AP performance, as well as identifying strategies to enhance instance detection, represents a promising direction for future research.

\subsection{Qualitative Results}
\cref{fig:quali} illustrates the qualitative results for the test segments of Bridge\,B (top row) and Bridge\,G (bottom row). Due to its lower performance, results from TopoCrack are excluded, as it frequently oversegments cracks, leading to a high number of false positives. Although nnU-Net demonstrates robust detection of wider cracks, it struggles to accurately segment narrower cracks and the tapering spurs of cracks.
Consistent with the quantitative findings, DetectionHMA exhibits superior performance, effectively capturing even the narrow sections of cracks. Unlike TopoCrack, both nnU-Net and DetectionHMA avoid oversegmenting cracks, resulting in a low number of false positives. However, for practical applications, it might be advantageous to increase the models' sensitivity. This could be achieved, for example, by downweighting the background predictions, thereby reducing type II errors. While this approach decreases the likelihood of missing a crack, it may come at the cost of generating more false positives.

The results for the test segment of Bridge\,G (\cref{fig:quali}, bottom) align with the quantitative findings regarding spalling and corrosion. nnU-Net excels in detecting corrosion, while DetectionHMA encounters difficulties with several instances. nnU-Net consistently segments corrosion across various views, whereas DetectionHMA shows greater variability depending on the viewpoint. This inconsistency in image-level predictions for DetectionHMA adversely affects the mapping process, leading to false negatives at certain points. Implementing a more sophisticated fusion method that emphasizes positive responses and/or accounts for viewing conditions could enhance the results. Further exploration into optimizing the contribution of different views offers a promising avenue for future research.

Regarding spalling detection, DetectionHMA struggles with the lower spalling instance in the zoomed view, whereas nnU-Net successfully identifies it. Although both methods detect portions of the spalling, they frequently fail to segment the entire extent. Considering that spalling usually leaves a distinct 3D footprint, leveraging native 3D detection could significantly improve the accuracy of spalling detection.

\section{Conclusion}
The evidence suggests that the proposed \enstrect{} is a highly effective tool for detecting structural damages in 2.5D space. Given the limited availability of 3D data, harnessing the strong performance of image-based models for damage segmentation emerges as a practical approach. The state-of-the-art models—TopoCrack, nnU-Net, and DetectionHMA—offer a solid foundation for mul\-ti-view damage detection. While TopoCrack tends to produce a high number of false positives, DetectionHMA excels in crack detection, and nnU-Net performs particularly well on areal damages such as spalling and corrosion. Segmentation performance can occasionally achieve an IoU of over 90\% at a positional tolerance of 4\,cm, but instance segmentation remains significantly lower and is not yet adequate for more advanced quantitative analyses.

Most research on damage detection concludes at the semantic segmentation stage. However, it is essential to emphasize that a comprehensive understanding of individual damage instances is necessary for further analytical processing, particularly regarding their size and extent. To achieve this, robust procedures for extracting the medial axes of cracks and the bounding polygons of areal damages are proposed. Point cloud contraction is employed to derive the medial axis of a crack, enabling the inference of both its length and, with additional processing, its width. For areal damages, principal component analysis (PCA) and alpha shapes are used to compute the bounding polygon, allowing for the accurate estimation of the damage's areal extent.

The general-purpose, self-configuring approach of nnU-Net demonstrated strong performance in detecting areal damages. However, it falls short of DetectionHMA in crack segmentation, as DetectionHMA was meticulously fine-tuned for cracks and other damages represented in the S2DS dataset. Improving nnU-Net's ability to handle fine structures like cracks could pave the way for a comprehensive, general-purpose approach to the semantic segmentation of structural damages with minimal manual intervention.

Furthermore, investigating the reasons behind the significantly lower performance in instance-level detection compared to segmentation could provide valuable insights and drive improvements in instance-level detection. Given the challenges associated with spalling detection, exploring native 3D detection for structural damages could offer substantial benefits. Additionally, implementing more sophisticated fusion schemes could enhance the transfer of information from 2D to 3D, helping to compensate for false positives or missed detections in image-level segmentation.

Throughout this work, a significant limitation became evident: the challenging scalability of point clouds. In real-world inspection scenarios, cracks as narrow as 0.3\,mm are relevant, requiring point clouds of extremely high density for accurate representation. Given the extensive areas involved in bridge inspections, these point clouds can easily contain millions or billions of points. Current technology does not readily support the processing of such large-scale point clouds. Consequently, it is crucial to evaluate whether point clouds are a viable representation for real-world inspection scenarios. Conceptually, a bridge might be more effectively represented as a textured mesh, where the texture provides high visual resolution, and the mesh captures a certain level of geometric detail.

\bibliographystyle{splncs04}
\bibliography{./export.bib}

\begin{thebibliography}{10}
\providecommand{\url}[1]{\texttt{#1}}
\providecommand{\urlprefix}{URL }
\providecommand{\doi}[1]{https://doi.org/#1}

\bibitem{badrinarayanan2017segnet}
Badrinarayanan, V., Kendall, A., Cipolla, R.: Segnet: A deep convolutional
  encoder-decoder architecture for image segmentation. IEEE Transactions on
  Pattern Analysis and Machine Intelligence  \textbf{39},  2481--2495 (2017).
  \doi{10.1109/TPAMI.2016.2644615}

\bibitem{bai2021detecting}
Bai, Y., Sezen, H., Yilmaz, A.: Detecting cracks and spalling automatically in
  extreme events by end-to-end deep learning frameworks. ISPRS Annals of the
  Photogrammetry, Remote Sensing and Spatial Information Sciences
  \textbf{V-2-2021},  161--168 (2021).
  \doi{10.5194/isprs-annals-V-2-2021-161-2021}

\bibitem{benz2023ai}
Benz, C., Debus, P., Gebhardt, T., Friedli, E., Schuhbäck, S., Zimmer, T.,
  Morgenthal, G., Rodehorst, V.: Ai-assisted inspection of concrete surfaces at
  dams. Allgemeine Vermessungs-Nachrichten (avn) (6)  (2023),
  \url{https://gispoint.de/artikelarchiv/avn/2023/avn-ausgabe-062023/7962-ki-gestuetzte-inspektion-von-betonoberflaechen-an-talsperren.html}

\bibitem{benz2019crack}
Benz, C., Debus, P., Ha, H.K., Rodehorst, V.: Crack segmentation on uas-based
  imagery using transfer learning. In: 2019 International Conference on Image
  and Vision Computing New Zealand (IVCNZ). pp.~1--6. IEEE (2019).
  \doi{10.1109/IVCNZ48456.2019.8960998}

\bibitem{benz2021model}
Benz, C., Rodehorst, V.: Model-based crack width estimation using rectangle
  transform. In: 2021 17th International Conference on Machine Vision and
  Applications (MVA). pp.~1--5. IEEE (2021).
  \doi{10.23919/MVA51890.2021.9511346}

\bibitem{benz2022image}
Benz, C., Rodehorst, V.: Image-based detection of structural defects using
  hierarchical multi-scale attention. In: Andres, B., Bernard, F., Cremers, D.,
  Frintrop, S., Goldlücke, B., Ihrke, I. (eds.) Pattern Recognition. DAGM GCPR
  2022. Lecture Notes in Computer Science, vol 13485. pp. 337--353. Springer
  (2022). \doi{10.1007/978-3-031-16788-1_21}

\bibitem{benz2024mvcrackvit}
Benz, C., Rodehorst, V.: Mvcrackvit: Robust multi-view crack detection for
  point cloud segmentation using view attention. In: 2024 IEEE International
  Conference on Image Processing (ICIP) (2024), (accepted manuscript)

\bibitem{benz2024omnicrack30k}
Benz, C., Rodehorst, V.: Omnicrack30k: A benchmark for crack segmentation and
  the reasonable effectiveness of transfer learning. In: Proceedings of the
  IEEE/CVF Conference on Computer Vision and Pattern Recognition (CVPR)
  Workshops. pp. 3876--3886 (2024),
  \url{https://openaccess.thecvf.com/content/CVPR2024W/VAND/html/Benz_OmniCrack30k_A_Benchmark_for_Crack_Segmentation_and_the_Reasonable_Effectiveness_CVPRW_2024_paper.html}

\bibitem{bergmann2022the}
Bergmann, P., Jin, X., Sattlegger, D., Steger, C.: The mvtec 3d-ad dataset for
  unsupervised 3d anomaly detection and localization. In: Proceedings of the
  17th International Joint Conference on Computer Vision, Imaging and Computer
  Graphics Theory and Applications (VISIGRAPP). pp. 202--213. SciTePress
  (2022). \doi{10.5220/0010865000003124}

\bibitem{bergmann2023anomaly}
Bergmann, P., Sattlegger, D.: Anomaly detection in 3d point clouds using deep
  geometric descriptors. In: 2023 IEEE/CVF Winter Conference on Applications of
  Computer Vision (WACV). pp. 2612--2622. IEEE (2023).
  \doi{10.1109/WACV56688.2023.00264}

\bibitem{bianchi2022development}
Bianchi, E., Hebdon, M.: Development of extendable open-source structural
  inspection datasets. Journal of Computing in Civil Engineering  \textbf{36}
  (2022). \doi{10.1061/(ASCE)CP.1943-5487.0001045}

\bibitem{bianchi2022visual}
Bianchi, E., Hebdon, M.: Visual structural inspection datasets. Automation in
  Construction  \textbf{139},  104299 (2022).
  \doi{10.1016/j.autcon.2022.104299}

\bibitem{cao2010point}
Cao, J., Tagliasacchi, A., Olson, M., Zhang, H., Su, Z.: Point cloud skeletons
  via laplacian based contraction. In: 2010 Shape Modeling International
  Conference. pp. 187--197. IEEE (2010). \doi{10.1109/SMI.2010.25}

\bibitem{cha2017deep}
Cha, Y., Choi, W., Büyüköztürk, O.: Deep learning‐based crack damage
  detection using convolutional neural networks. Computer-Aided Civil and
  Infrastructure Engineering  \textbf{32},  361--378 (2017).
  \doi{10.1111/mice.12263}

\bibitem{chen2018nb}
Chen, F.C., Jahanshahi, M.R.: Nb-cnn: Deep learning-based crack detection using
  convolutional neural network and naïve bayes data fusion. IEEE Transactions
  on Industrial Electronics  \textbf{65},  4392--4400 (2018).
  \doi{10.1109/TIE.2017.2764844}

\bibitem{chen2022crackembed}
Chen, J., Cho, Y.K.: Crackembed: Point feature embedding for crack segmentation
  from disaster site point clouds with anomaly detection. Advanced Engineering
  Informatics  \textbf{52},  101550 (2022). \doi{10.1016/j.aei.2022.101550}

\bibitem{chen2018encoder}
Chen, L.C., Zhu, Y., Papandreou, G., Schroff, F., Adam, H.: Encoder-decoder
  with atrous separable convolution for semantic image segmentation. In:
  Ferrari, V., Hebert, M., Sminchisescu, C., Weiss, Y. (eds.) Computer Vision
  – ECCV 2018. pp. 833--851. Springer (2018).
  \doi{10.1007/978-3-030-01234-2_49}

\bibitem{chen2023the}
Chen, Z., Zhang, J., Lai, Z., Zhu, G., Liu, Z., Chen, J., Li, J.: The devil is
  in the crack orientation: A new perspective for crack detection. In: 2023
  IEEE/CVF International Conference on Computer Vision (ICCV). pp. 6630--6640.
  IEEE (2023). \doi{10.1109/ICCV51070.2023.00612}

\bibitem{cheng2022masked}
Cheng, B., Misra, I., Schwing, A.G., Kirillov, A., Girdhar, R.:
  Masked-attention mask transformer for universal image segmentation. In: 2022
  IEEE/CVF Conference on Computer Vision and Pattern Recognition (CVPR). pp.
  1280--1289. IEEE (2022). \doi{10.1109/CVPR52688.2022.00135}

\bibitem{dorafshan2018comparison}
Dorafshan, S., Thomas, R.J., Maguire, M.: Comparison of deep convolutional
  neural networks and edge detectors for image-based crack detection in
  concrete. Construction and Building Materials  \textbf{186},  1031--1045
  (2018). \doi{10.1016/j.conbuildmat.2018.08.011}

\bibitem{dorafshan2018sdnet2018}
Dorafshan, S., Thomas, R.J., Maguire, M.: Sdnet2018: An annotated image dataset
  for non-contact concrete crack detection using deep convolutional neural
  networks. Data in Brief  \textbf{21},  1664--1668 (2018).
  \doi{10.1016/j.dib.2018.11.015}

\bibitem{edelsbrunner1983on}
Edelsbrunner, H., Kirkpatrick, D., Seidel, R.: On the shape of a set of points
  in the plane. IEEE Transactions on Information Theory  \textbf{29},  551--559
  (1983). \doi{10.1109/TIT.1983.1056714}

\bibitem{edelsbrunner1994three}
Edelsbrunner, H., Mücke, E.P.: Three-dimensional alpha shapes. ACM
  Transactions on Graphics  \textbf{13},  43--72 (1994).
  \doi{10.1145/174462.156635}

\bibitem{ester1996a}
Ester, M., Kriegel, H.P., Sander, J., Xu, X.: A density-based algorithm for
  discovering clusters in large spatial databases with noise. In: Proceedings
  of the International Conference on Knowledge Discovery and Data Mining (KDD).
  pp. 226--231 (1996), \url{https://dl.acm.org/doi/10.5555/3001460.3001507}

\bibitem{fang2024eva}
Fang, Y., Sun, Q., Wang, X., Huang, T., Wang, X., Cao, Y.: Eva-02: A visual
  representation for neon genesis. Image and Vision Computing  \textbf{149},
  105171 (2024). \doi{10.1016/j.imavis.2024.105171}

\bibitem{flotzinger2024dacl}
Flotzinger, J., Rösch, P.J., Benz, C., Ahmad, M., Cankaya, M., Mayer, H.,
  Rodehorst, V., Oswald, N., Braml, T.: dacl-challenge: Semantic segmentation
  during visual bridge inspections. In: 2024 IEEE/CVF Winter Conference on
  Applications of Computer Vision Workshops (WACVW). pp. 716--725. IEEE (2024).
  \doi{10.1109/WACVW60836.2024.00084}

\bibitem{flotzinger2024dacl10k}
Flotzinger, J., Rösch, P.J., Braml, T.: dacl10k: Benchmark for semantic bridge
  damage segmentation. In: 2024 IEEE/CVF Winter Conference on Applications of
  Computer Vision (WACV). pp. 8611--8620. IEEE (2024).
  \doi{10.1109/WACV57701.2024.00843}

\bibitem{flotzinger2022building}
Flotzinger, J., Rösch, P.J., Oswald, N., Braml, T.: Building inspection
  toolkit: Unified evaluation and strong baselines for bridge damage
  recognition. In: 2022 IEEE International Conference on Image Processing
  (ICIP). pp. 1221--1225. IEEE (2022). \doi{10.1109/ICIP46576.2022.9897743}

\bibitem{gkioxari2019mesh}
Gkioxari, G., Johnson, J., Malik, J.: Mesh r-cnn. In: 2019 IEEE/CVF
  International Conference on Computer Vision (ICCV). pp. 9784--9794. IEEE
  (2019). \doi{10.1109/ICCV.2019.00988}

\bibitem{he2013guided}
He, K., Sun, J., Tang, X.: Guided image filtering. IEEE Transactions on Pattern
  Analysis and Machine Intelligence  \textbf{35},  1397--1409 (2013).
  \doi{10.1109/TPAMI.2012.213}

\bibitem{huang2017densely}
Huang, G., Liu, Z., Maaten, L.V.D., Weinberger, K.Q.: Densely connected
  convolutional networks. In: 2017 IEEE Conference on Computer Vision and
  Pattern Recognition (CVPR). pp. 2261--2269. IEEE (2017).
  \doi{10.1109/CVPR.2017.243}

\bibitem{huang2014a}
Huang, J., Liu, W., Sun, X.: A pavement crack detection method combining 2d
  with 3d information based on dempster‐shafer theory. Computer-Aided Civil
  and Infrastructure Engineering  \textbf{29},  299--313 (2014).
  \doi{10.1111/mice.12041}

\bibitem{iglovikov2018ternausnet}
Iglovikov, V., Shvets, A.: Ternausnet: U-net with vgg11 encoder pre-trained on
  imagenet for image segmentation. arXiv preprint arXiv:1801.05746  (2018).
  \doi{10.48550/arXiv.1801.05746}

\bibitem{isensee2021nnu}
Isensee, F., Jaeger, P.F., Kohl, S.A.A., Petersen, J., Maier-Hein, K.H.:
  nnu-net: a self-configuring method for deep learning-based biomedical image
  segmentation. Nature Methods  \textbf{18},  203--211 (2021).
  \doi{10.1038/s41592-020-01008-z}

\bibitem{jahanshahi2012adaptive}
Jahanshahi, M.R., Masri, S.F.: Adaptive vision-based crack detection using 3d
  scene reconstruction for condition assessment of structures. Automation in
  Construction  \textbf{22},  567--576 (2012).
  \doi{10.1016/j.autcon.2011.11.018}

\bibitem{knapitsch2017tanks}
Knapitsch, A., Park, J., Zhou, Q.Y., Koltun, V.: Tanks and temples. ACM
  Transactions on Graphics  \textbf{36},  1--13 (2017).
  \doi{10.1145/3072959.3073599}

\bibitem{krizhevsky2012imagenet}
Krizhevsky, A., Sutskever, I., Hinton, G.E.: Imagenet classification with deep
  convolutional neural networks. In: Pereira, F., Burges, C., Bottou, L.,
  Weinberger, K. (eds.) Advances in Neural Information Processing Systems 25.
  pp. 1097--1105. Curran Associates, Inc. (2012),
  \url{https://papers.nips.cc/paper_files/paper/2012/hash/c399862d3b9d6b76c8436e924a68c45b-Abstract.html}

\bibitem{kulkarni2022crackseg9k}
Kulkarni, S., Singh, S., Balakrishnan, D., Sharma, S., Devunuri, S., Korlapati,
  S.C.R.: Crackseg9k: A collection and benchmark for crack segmentation
  datasets and frameworks. In: Karlinsky, L., Michaeli, T., Nishino, K. (eds.)
  Computer Vision -- ECCV 2022 Workshops. pp. 179--195. Springer (2023).
  \doi{10.1007/978-3-031-25082-8_12}

\bibitem{lee2015deeply}
Lee, C.Y., Xie, S., Gallagher, P., Zhang, Z., Tu, Z.: Deeply-supervised nets.
  Artificial Intelligence and Statistics  \textbf{38},  562--570 (2015),
  \url{https://proceedings.mlr.press/v38/lee15a.html}

\bibitem{lin2017focal}
Lin, T.Y., Goyal, P., Girshick, R., He, K., Dollar, P.: Focal loss for dense
  object detection. In: 2017 IEEE International Conference on Computer Vision
  (ICCV). pp. 2999--3007. IEEE (2017). \doi{10.1109/ICCV.2017.324}

\bibitem{liu2021crackformer}
Liu, H., Miao, X., Mertz, C., Xu, C., Kong, H.: Crackformer: Transformer
  network for fine-grained crack detection. In: 2021 IEEE/CVF International
  Conference on Computer Vision (ICCV). pp. 3763--3772. IEEE (2021).
  \doi{10.1109/ICCV48922.2021.00376}

\bibitem{liu2019deepcrack}
Liu, Y., Yao, J., Lu, X., Xie, R., Li, L.: Deepcrack: A deep hierarchical
  feature learning architecture for crack segmentation. Neurocomputing
  \textbf{338},  139--153 (2019). \doi{10.1016/j.neucom.2019.01.036}

\bibitem{liu2019computer}
Liu, Z., Cao, Y., Wang, Y., Wang, W.: Computer vision-based concrete crack
  detection using u-net fully convolutional networks. Automation in
  Construction  \textbf{104},  129--139 (2019).
  \doi{10.1016/j.autcon.2019.04.005}

\bibitem{liu2022a}
Liu, Z., Mao, H., Wu, C.Y., Feichtenhofer, C., Darrell, T., Xie, S.: A convnet
  for the 2020s. In: 2022 IEEE/CVF Conference on Computer Vision and Pattern
  Recognition (CVPR). pp. 11966--11976. IEEE (2022).
  \doi{10.1109/CVPR52688.2022.01167}

\bibitem{long2015fully}
Long, J., Shelhamer, E., Darrell, T.: Fully convolutional networks for semantic
  segmentation. In: 2015 IEEE Conference on Computer Vision and Pattern
  Recognition (CVPR). pp. 3431--3440. IEEE (2015).
  \doi{10.1109/CVPR.2015.7298965}

\bibitem{mundt2019meta}
Mundt, M., Majumder, S., Murali, S., Panetsos, P., Ramesh, V.: Meta-learning
  convolutional neural architectures for multi-target concrete defect
  classification with the concrete defect bridge image dataset. In: 2019
  IEEE/CVF Conference on Computer Vision and Pattern Recognition (CVPR). pp.
  11188--11197. IEEE (2019). \doi{10.1109/CVPR.2019.01145}

\bibitem{pantoja2023damage}
Pantoja-Rosero, B., Achanta, R., Beyer, K.: Damage-augmented digital twins
  towards the automated inspection of buildings. Automation in Construction
  \textbf{150},  104842 (2023). \doi{10.1016/j.autcon.2023.104842}

\bibitem{pantoja2022topo}
Pantoja-Rosero, B., Oner, D., Kozinski, M., Achanta, R., Fua, P., Perez-Cruz,
  F., Beyer, K.: Topo-loss for continuity-preserving crack detection using deep
  learning. Construction and Building Materials  \textbf{344},  128264 (2022).
  \doi{10.1016/j.conbuildmat.2022.128264}

\bibitem{ronneberger2015u}
Ronneberger, O., Fischer, P., Brox, T.: U-net: Convolutional networks for
  biomedical image segmentation. In: Navab, N., Hornegger, J., Wells, W.M.,
  Frangi, A.F. (eds.) Medical Image Computing and Computer-Assisted
  Intervention -- MICCAI 2015. pp. 234--241. Springer (2015).
  \doi{10.1007/978-3-319-24574-4_28}

\bibitem{schonberger2016structure}
Schönberger, J.L., Frahm, J.M.: Structure-from-motion revisited. In: 2016 IEEE
  Conference on Computer Vision and Pattern Recognition (CVPR). pp. 4104--4113.
  IEEE (2016). \doi{10.1109/CVPR.2016.445}

\bibitem{schonberger2016pixelwise}
Schönberger, J.L., Zheng, E., Frahm, J.M., Pollefeys, M.: Pixelwise view
  selection for unstructured multi-view stereo. In: Leibe, B., Matas, J., Sebe,
  N., Welling, M. (eds.) Computer Vision -- ECCV 2016. pp. 501--518. Springer
  (2016). \doi{10.1007/978-3-319-46487-9_31}

\bibitem{simonyan2015very}
Simonyan, K., Zisserman, A.: Very deep convolutional networks for large-scale
  image recognition. In: 3rd International Conference on Learning
  Representations (ICLR). pp. 1--14. Computational and Biological Learning
  Society (2015). \doi{10.48550/arXiv.1409.1556}

\bibitem{tao2020hierarchical}
Tao, A., Sapra, K., Catanzaro, B.: Hierarchical multi-scale attention for
  semantic segmentation. arXiv preprint arXiv:2005.10821  (2020).
  \doi{10.48550/arXiv.2005.10821}

\bibitem{torok2014image}
Torok, M.M., Golparvar-Fard, M., Kochersberger, K.B.: Image-based automated 3d
  crack detection for post-disaster building assessment. Journal of Computing
  in Civil Engineering  \textbf{28} (2014).
  \doi{10.1061/(ASCE)CP.1943-5487.0000334}

\bibitem{wang2018pixel2mesh}
Wang, N., Zhang, Y., Li, Z., Fu, Y., Liu, W., Jiang, Y.G.: Pixel2mesh:
  Generating 3d mesh models from single rgb images. In: Ferrari, V., Hebert,
  M., Sminchisescu, C., Weiss, Y. (eds.) Computer Vision -- ECCV 2018. pp.
  55--71. Springer (2018). \doi{10.1007/978-3-030-01252-6_4}

\bibitem{xie2015holistically}
Xie, S., Tu, Z.: Holistically-nested edge detection. In: 2015 IEEE
  International Conference on Computer Vision (ICCV). pp. 1395--1403. IEEE
  (2015). \doi{10.1109/ICCV.2015.164}

\bibitem{yang2020feature}
Yang, F., Zhang, L., Yu, S., Prokhorov, D., Mei, X., Ling, H.: Feature pyramid
  and hierarchical boosting network for pavement crack detection. IEEE
  Transactions on Intelligent Transportation Systems  \textbf{21},  1525--1535
  (2020). \doi{10.1109/TITS.2019.2910595}

\bibitem{yang2017a}
Yang, L., Li, B., Li, W., Liu, Z., Yang, G., Xiao, J.: A robotic system towards
  concrete structure spalling and crack database. In: 2017 IEEE International
  Conference on Robotics and Biomimetics (ROBIO). pp. 1276--1281. IEEE (2017).
  \doi{10.1109/ROBIO.2017.8324593}

\bibitem{yang2018automatic}
Yang, X., Li, H., Yu, Y., Luo, X., Huang, T., Yang, X.: Automatic pixel‐level
  crack detection and measurement using fully convolutional network.
  Computer-Aided Civil and Infrastructure Engineering  \textbf{33},  1090--1109
  (2018). \doi{10.1111/mice.12412}

\bibitem{yuan2020object}
Yuan, Y., Chen, X., Wang, J.: Object-contextual representations for semantic
  segmentation. In: Vedaldi, A., Bischof, H., Brox, T., Frahm, J.M. (eds.)
  Computer Vision -- ECCV 2020. pp. 173--190. Springer (2020).
  \doi{10.1007/978-3-030-58539-6_11}

\bibitem{zhang2018deep}
Zhang, A., Wang, K.C.P., Fei, Y., Liu, Y., Tao, S., Chen, C., Li, J.Q., Li, B.:
  Deep learning–based fully automated pavement crack detection on 3d asphalt
  surfaces with an improved cracknet. Journal of Computing in Civil Engineering
   \textbf{32} (2018). \doi{10.1061/(ASCE)CP.1943-5487.0000775}

\bibitem{zhang2017automated}
Zhang, A., Wang, K.C.P., Li, B., Yang, E., Dai, X., Peng, Y., Fei, Y., Liu, Y.,
  Li, J.Q., Chen, C.: Automated pixel‐level pavement crack detection on 3d
  asphalt surfaces using a deep‐learning network. Computer-Aided Civil and
  Infrastructure Engineering  \textbf{32},  805--819 (2017).
  \doi{10.1111/mice.12297}

\bibitem{zhang2016road}
Zhang, L., Yang, F., Zhang, Y.D., Zhu, Y.J.: Road crack detection using deep
  convolutional neural network. In: 2016 IEEE International Conference on Image
  Processing (ICIP). pp. 3708--3712. IEEE (2016).
  \doi{10.1109/ICIP.2016.7533052}

\bibitem{zhao2019region}
Zhao, S., Wang, Y., Yang, Z., Cai, D.: Region mutual information loss for
  semantic segmentation. In: Wallach, H., Larochelle, H., Beygelzimer, A.,
  d'Alché Buc, F., Fox, E., Garnett, R. (eds.) Advances in Neural Information
  Processing Systems 32. Curran Associates, Inc. (2019),
  \url{https://papers.nips.cc/paper_files/paper/2019/hash/a67c8c9a961b4182688768dd9ba015fe-Abstract.html}

\bibitem{zheng2015conditional}
Zheng, S., Jayasumana, S., Romera-Paredes, B., Vineet, V., Su, Z., Du, D.,
  Huang, C., Torr, P.H.S.: Conditional random fields as recurrent neural
  networks. In: 2015 IEEE International Conference on Computer Vision (ICCV).
  pp. 1529--1537. IEEE (2015). \doi{10.1109/ICCV.2015.179}

\bibitem{zou2019deepcrack}
Zou, Q., Zhang, Z., Li, Q., Qi, X., Wang, Q., Wang, S.: Deepcrack: Learning
  hierarchical convolutional features for crack detection. IEEE Transactions on
  Image Processing  \textbf{28},  1498--1512 (2019).
  \doi{10.1109/TIP.2018.2878966}

\end{thebibliography}
\end{document}